\def\GASPArxiv{}
\documentclass[letterpaper, 10 pt, conference]{ieeeconf}  

\IEEEoverridecommandlockouts                              

\usepackage{cite}
\usepackage{pgfplots}

\usepgfplotslibrary{fillbetween}
\usepackage{amsmath,amssymb,amsfonts}
\usepackage{algorithmic}
\usepackage{graphicx}
\usepackage{textcomp}
\usepackage{hyperref}
\usepackage{placeins}
\usepackage{xcolor}
\usepackage{booktabs}
\usepackage{multirow}
\def\BibTeX{{\rm B\kern-.05em{\sc i\kern-.025em b}\kern-.08em
    T\kern-.1667em\lower.7ex\hbox{E}\kern-.125emX}}

\usepackage{tikz}
\usepackage{pgfplots}
\usepgfplotslibrary{groupplots}
\pgfplotsset{compat=1.18}
\usepackage{amsmath, amssymb}
\usetikzlibrary{positioning, shapes.geometric, arrows.meta, calc, fit, backgrounds}
\pgfplotsset{compat=1.18}

\usepackage{caption}
\usepackage{xspace}
\usepackage{tabularx}
\usepackage{array}
\newif\ifanonymized
\anonymizedfalse
\ifdefined\GASPAnonymized
  \anonymizedtrue
\fi

\newif\ifarxiv
\arxivfalse
\ifdefined\GASPArxiv
  \arxivtrue
\fi

\newif\ifshowtodos
\showtodosfalse
\ifdefined\GASPShowTodos
  \showtodostrue
\fi

\ifanonymized
  \hypersetup{
    pdftitle={GASP: GPU-Accelerated Safe Planner for Real-Time Collision-Aware Motion Generation with Latent Trajectory Sampling},
    pdfauthor={},
    pdfsubject={},
    pdfkeywords={}
  }
\else
  \hypersetup{
    pdftitle={GASP: GPU-Accelerated Safe Planner for Real-Time Collision-Aware Motion Generation with Latent Trajectory Sampling},
    pdfauthor={Colin Merk, Stefanos Charalambous, Peter Durr, Farshad Khadivar},
    pdfsubject={},
    pdfkeywords={Motion and Path Planning, Collision Avoidance, Constrained Motion Planning, Machine Learning for Robot Control}
  }
\fi

\newcounter{todocounter}

\ifshowtodos
  \newcommand{\todo}[1]{%
    \stepcounter{todocounter}%
    \textcolor{red}{{TODO\thetodocounter:} #1}%
  }
\else
  \newcommand{\todo}[1]{}%
\fi
\newcounter{questioncounter}

\begin{document}

\title{\LARGE \bf
GASP: GPU-Accelerated Safe Planner for Real-Time Collision-Aware Motion Generation with Latent Trajectory Sampling
}

\ifanonymized
\author{Anonymous Authors}
\else
\author{%
Colin Merk$^{1}$,
Stefanos Charalambous$^{1}$,
Peter D\"urr$^{1}$,
Farshad Khadivar$^{1}$%
\thanks{$^{1}$Sony, Zurich, Switzerland.
        {\tt\small
        colin.robin.merk@gmail.com,
        stefchr.cy@gmail.com,
        peter.duerr@ieee.org,
        fr.khadivar@gmail.com}}
\thanks{Corresponding author: Colin Merk (colin.robin.merk@gmail.com).}
\ifarxiv
\thanks{This work has been submitted to the IEEE for possible publication.
Copyright may be transferred without notice, after which this version may no longer be accessible.}
\fi
}
\fi

\maketitle

\begin{abstract}

We present GASP, a GPU-Accelerated Safe Planner for real-time, collision-aware joint-space motion generation in known environments.
GASP combines a clamped B-spline trajectory parameterization with a convolutional residual neural network that predicts the free interior control points, while analytically inserted boundary control points enforce initial and final derivative constraints for collision-aware planning under non-stationary conditions. A conditional variational autoencoder samples multiple trajectory candidates, which are decoded and validated in parallel on the GPU, yielding a batched planner for collision-aware coupled joint-space motion with near-millisecond inference. We validate GASP as an online motion-generation module, where it achieves analytical-level success rates with high collision-aware feasibility and substantially reduces inference time relative to GPU-based trajectory optimization. We further deploy GASP as a reinforcement-learning reset planner in competitive robotic table tennis, matching the baseline return rate while roughly halving training-time collisions.
\end{abstract}

\begin{keywords}
Motion and Path Planning, Collision Avoidance, Constrained Motion Planning, Machine Learning for Robot Control.
\end{keywords}

\section{Introduction}
Closed-loop robotic manipulation requires real-time motion planners that repeatedly generate collision-aware and dynamically feasible joint-space trajectories under changing boundary conditions caused by moving targets, perception updates, model uncertainty, and tracking errors. In practical settings such as repeated manipulation in known static environments \cite{ni2023progressive, kicki2023fastkinodynamicplanningconstraint}, the planner must maintain high-frequency, deterministic inference while satisfying safety and feasibility constraints. This remains challenging for optimization-based planners \cite{huang2024diffusionseederseedingmotionoptimization}, whose computational cost and runtime variability can limit their use inside online learning and control systems. The constraint is especially relevant in GPU-based reinforcement-learning pipelines \cite{rudin2022learningwalkminutesusing}, where running planning on the same hardware can simplify integration and reduce latency.

\begin{figure}[t]
\centering
\includegraphics[width=0.9\columnwidth,trim={2.5cm 2.5cm 6cm 2cm},clip]{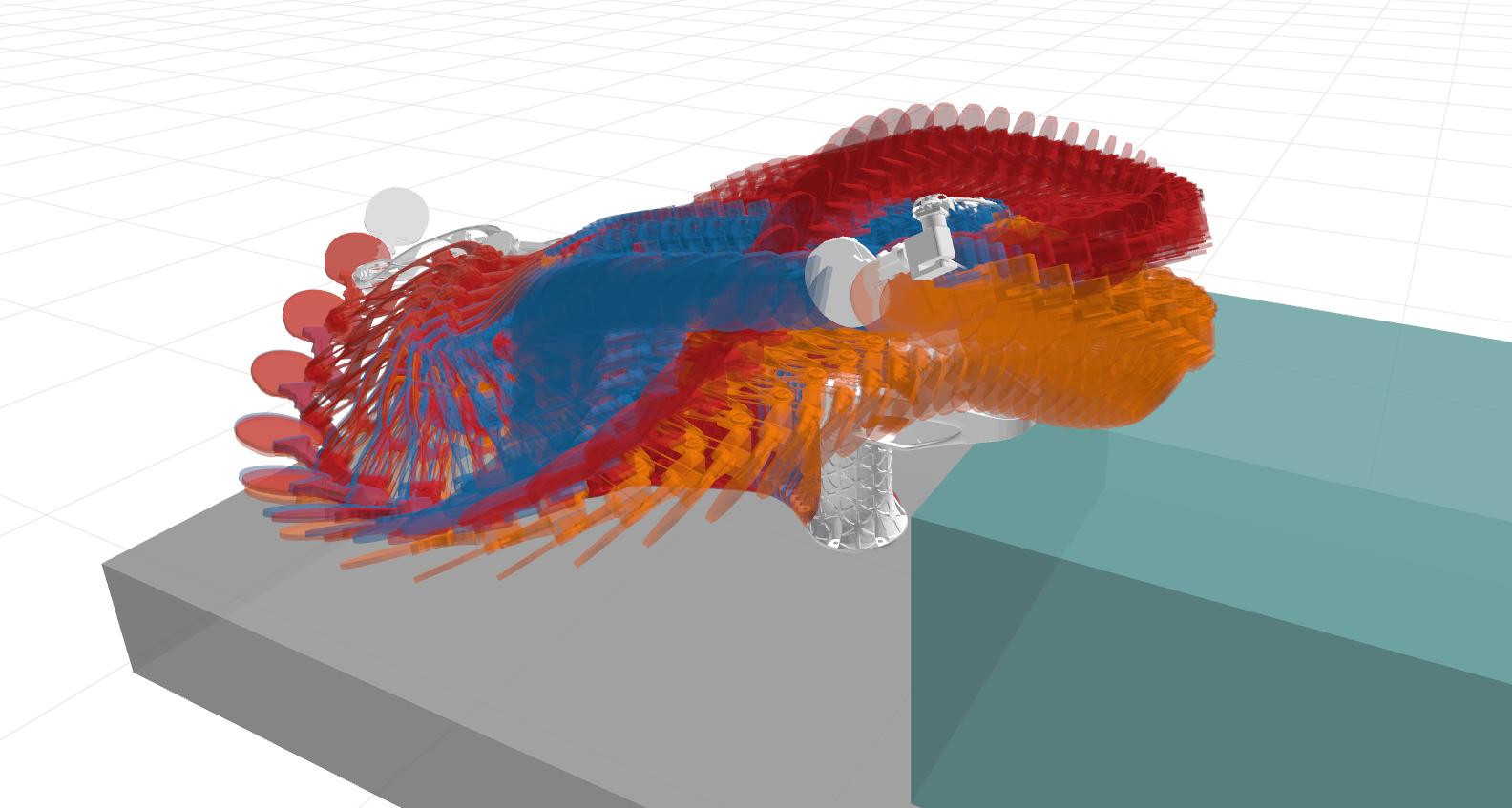}
\caption{
Candidate trajectories sampled by GASP for a single start--goal query.
Batched GPU inference predicts multiple sets of interior B-spline control points, while boundary control points are inserted analytically from the prescribed endpoint conditions.
All candidates are evaluated in parallel for joint limits, dynamic limits, self-collision, and workspace collisions.
}
\label{fig:gpu_safe_planner_overview}
\vspace{-5mm}
\end{figure}

\begin{figure*}[t]
    \centering
    \includegraphics[width=0.97\linewidth]{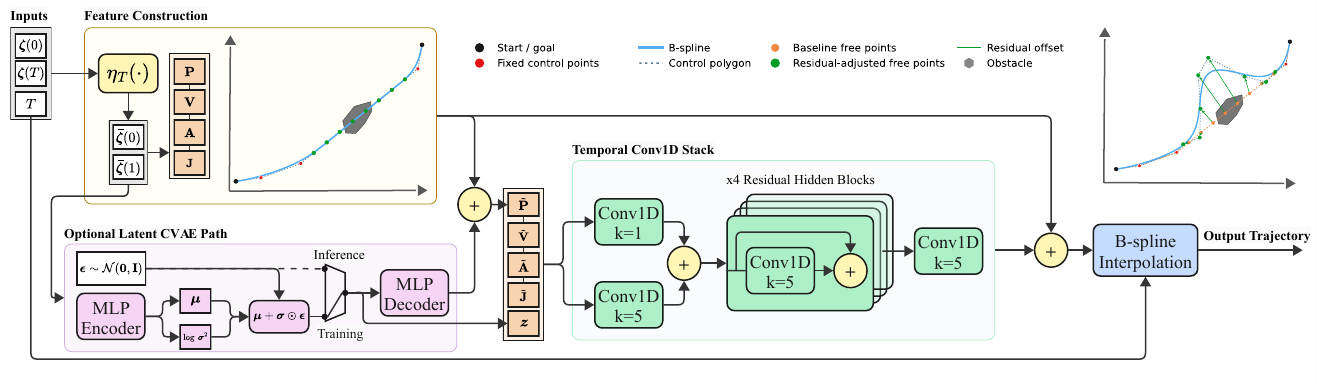}
\caption{Overview of the GASP pipeline. The model takes the boundary conditions $\zeta(0)$ and $\zeta(T)$, together with the trajectory horizon $T$, as inputs. The time-normalized start and goal states define fixed red B-spline control points encoding boundary velocity and acceleration constraints, while the free points are linearly interpolated to form the nominal blue trajectory. This trajectory is encoded by kinematic descriptors $({\mathbf{P}}, \mathbf{V}, \mathbf{A}, \mathbf{J})$. An optional CVAE branch models trajectory variations by encoding $(\boldsymbol{\mu}, \log \boldsymbol{\sigma}^2)$ during training and sampling $\mathbf{z} \sim \mathcal{N}(\mathbf{0}, \mathbf{I})$ at inference. Decoded latent residuals are concatenated with the kinematic features and passed to a network with parallel $k\in\{1,5\}$ 1D convolutions, four residual blocks, and a final $k=5$ convolution, which predicts residual offsets that refine the free control points while preserving the fixed boundary points. The resulting adjusted control points are mapped to the dense B-spline trajectory via matrix multiplication. In all experiments, GASP uses a custom-built robotic arm with 8 degrees of freedom, a degree-7 B-spline with 13 control points, 8 free control points, initial position, velocity, and acceleration and final position, and velocity boundary conditions, and a five-layer convolutional residual predictor with 512 hidden channels, kernel size 5, and a 2D latent space.
}
\label{fig:pipeline}
\end{figure*}
This motivates compact trajectory representations that encode smooth motion and support parallel feasibility evaluation. Spline-based parameterizations \cite{kicki2024kinodynamic} provide analytical derivatives and efficient dense evaluation, making B-splines attractive for optimization-based and learned motion planners \cite{sundaralingam2026curobov2dynamicsawaremotiongeneration, kicki2023fastkinodynamicplanningconstraint}. They are also well suited for boundary-conditioned planning: recent learned kinodynamic planners enforce endpoint constraints analytically while predicting the remaining free trajectory parameters \cite{kicki2023fastkinodynamicplanningconstraint}, reducing the learning burden while preserving boundary feasibility. Other approaches further learn temporal scaling or adaptive spline parameterizations to handle varying motion durations \cite{kicki2022speeding}.

Here, we present GASP, a GPU-Accelerated Safe Planner for real-time joint-space planning in known static environments. GASP uses a learned generator with a boundary-conditioned clamped B-spline representation: the network predicts only unconstrained interior control-point residuals, while the boundary control points required to satisfy endpoint derivative constraints are computed analytically. This reduces prediction dimensionality, guarantees the prescribed boundary conditions, and produces smooth trajectories by construction.

GASP is designed for online deployment: trajectory sampling, dense spline evaluation, feasibility checking, and candidate selection are fully batched on the GPU. A lightweight conditional convolutional residual predictor samples multiple candidates for each start--goal query, and training uses differentiable feasibility penalties for joint limits, dynamic limits, self-collision, world-collision, and smoothness, without requiring expert trajectory supervision. The specific contributions of this work are:
\begin{itemize}
    \item A boundary-conditioned clamped B-spline planner that analytically enforces endpoint derivative constraints while learning only interior control-point residuals.
    \item A GPU-batched, sampleable neural planning pipeline that generates, decodes, collision-checks, and selects among multiple trajectory candidates in near-millisecond inference time.
    \item A feasibility-driven training and evaluation study showing collision-aware joint-space planning without expert trajectory supervision, with validation against Ruckig, cuRobo, neural baselines, and deployment in robotic table tennis.
\end{itemize}

We evaluate GASP against analytical trajectory generation and GPU-accelerated optimization-based planning. Specifically, we compare against Ruckig \cite{berscheid2021jerk}, which computes analytical time-optimal trajectories under kinematic and dynamic constraints but does not account for collisions, and cuRobo \cite{sundaralingam2023curoboparallelizedcollisionfreeminimumjerk}, which performs collision-aware trajectory optimization on the GPU. Across diverse planning datasets, GASP achieves success rates comparable to or higher than analytical solutions while adding collision-aware feasibility through parallel candidate validation. Compared with cuRobo, GASP achieves substantially higher success rates under non-stationary boundary conditions while reducing inference time, positioning it between the speed of analytical generation and the safety awareness of collision-aware optimization.

\section{Related Work}

Optimization-based planners formulate trajectory generation as constrained optimization over robot states. CHOMP~\cite{chomp} generates smooth, collision-free trajectories through iterative gradient updates, while GPU planners such as cuRobo~\cite{sundaralingam2023curoboparallelizedcollisionfreeminimumjerk} accelerate collision checking and trajectory optimization for replanning in static environments; related reactive systems combine GPU planning with perception for online motion generation~\cite{11457603}. These methods offer flexible constraint handling, but iterative optimization can introduce runtime variance and computational cost under rapidly changing boundary conditions. In contrast, analytical online trajectory generators such as Ruckig~\cite{berscheid2021jerk} compute jerk-limited trajectories with low latency and deterministic runtime, but typically ignore workspace collisions and coupled geometric feasibility.

Learning-based motion planners amortize trajectory generation through neural inference. Earlier methods accelerate sampling-based or optimization-based planning by predicting trajectory initializations or feasible motion priors \cite{kicki2022speeding}. Subsequent work extended this idea to direct kinodynamic trajectory generation in joint space: \cite{kicki2023fastkinodynamicplanningconstraint} predicts feasible trajectories under kinematic constraints at high inference speed and demonstrates stability inside a reinforcement-learning training loop. Progressive generation strategies have also been explored for constrained motion planning \cite{ni2023progressive}, refining trajectories incrementally to improve feasibility and stability.

Recent generative planners formulate motion planning as conditional trajectory sampling. Diffusion and flow-based models generate trajectory distributions rather than single deterministic solutions, improving multimodal coverage and robustness. Diffusion-based robot motion generators \cite{carvalho2025motion} show that stochastic samplers can represent diverse feasible trajectories while maintaining smoothness and constraint satisfaction, and flow-matching formulations have recently been explored for robotic trajectory generation and optimization \cite{mcallister2025flow}. These methods improve expressiveness, but their sampling procedures remain costly for sub-millisecond inference.

GASP is closest to learned kinodynamic trajectory generators that seek analytical-planner speed with optimization-style feasibility awareness. Unlike optimization-heavy pipelines, GASP predicts trajectories in a single forward pass and evaluates multiple candidates in parallel on the GPU. Unlike analytical generators such as Ruckig, it explicitly checks self-collision and workspace-collision feasibility. Compared with recent generative planners, GASP uses a lightweight residual B-spline parameterization with analytically enforced boundary constraints, enabling near-millisecond batched inference while remaining sampleable through a conditional latent variable.

\section{Method}
\label{sec:method}

Given initial and goal boundary states, and a horizon $T$, GASP predicts the unconstrained interior control points of a high-order B-spline using a convolutional residual network with a sampleable latent variable. These points are combined with analytically computed boundary control points and decoded into dense position, velocity, acceleration, and jerk trajectories. The overall pipeline is shown in Fig.~\ref{fig:pipeline}; implementation details are provided in Appendix~\ref{app:methods}.

\subsection{Problem Formulation}
\label{sec:problem_formulation}

Let $\mathbf{q}(t) \in \mathbb{R}^n$ denote the robot joint positions at physical time $t$, with velocity $\dot{\mathbf q}(t)$ and acceleration $\ddot{\mathbf q}(t)$. We define the trajectory state as $\zeta(t) = \left(\mathbf{q}(t), \dot{\mathbf q}(t), \ddot{\mathbf q}(t)\right)$, so a trajectory over horizon $T > 0$ is a map:

\begin{equation}
\label{eq:trajectory_map}
\zeta : [0,T] \rightarrow \mathbb{R}^{3n}.
\end{equation}

The planning problem is to generate a smooth, collision-free, and kinodynamically feasible trajectory that transitions from an initial boundary state to a desired final boundary state. For the formulation considered here, we assume that these boundary states are
\begin{equation}
\label{eq:boundary_states}
\zeta(0)
=
(\mathbf{q}_0, \dot{\mathbf q}_0, \ddot{\mathbf q}_0),
\qquad
\zeta(T)
=
(\mathbf{q}_T, \dot{\mathbf q}_T).
\end{equation}

Thus, the initial state is constrained up to acceleration and the final state up to velocity. The planner must satisfy the boundary conditions in Eq.~\ref{eq:boundary_states}, the joint-limit constraints in Eq.~\ref{eq:bounds}, and collision avoidance for all $t \in [0,T]$.

\subsubsection{Joint limits}

Let the admissible joint-limit intervals be

\begin{equation*}
\begin{aligned}
    \mathcal{L}_{\mathbf q} &= [\mathbf q_{l}, \mathbf q_{u}], &
    \mathcal{L}_{\dot{\mathbf q}} &= [\dot{\mathbf q}_{l}, \dot{\mathbf q}_{u}], \\
    \mathcal{L}_{\ddot{\mathbf q}} &= [\ddot{\mathbf q}_{l}, \ddot{\mathbf q}_{u}], &
    \mathcal{L}_{\dddot{\mathbf q}} &= [\dddot{\mathbf q}_{l}, \dddot{\mathbf q}_{u}],
\end{aligned}
\end{equation*}

where $l$ and $u$ denote lower and upper bounds. A trajectory is joint-limit feasible if, for every sampled time $t_k$,

\begin{equation}
\begin{aligned}
\label{eq:bounds}
    \mathbf q(t_k) &\in \mathcal{L}_{\mathbf q}, & \dot{\mathbf q}(t_k) \in \mathcal{L}_{\dot{\mathbf q}}, \\
    \ddot{\mathbf q}(t_k) &\in \mathcal{L}_{\ddot{\mathbf q}}, &\dddot{\mathbf q}(t_k) \in \mathcal{L}_{\dddot{\mathbf q}}.
\end{aligned}
\end{equation}

\subsubsection{Collision avoidance}

\begin{figure}[t]
\centering
\includegraphics[width=0.7\columnwidth,trim={10cm 12cm 19cm 8cm},clip]{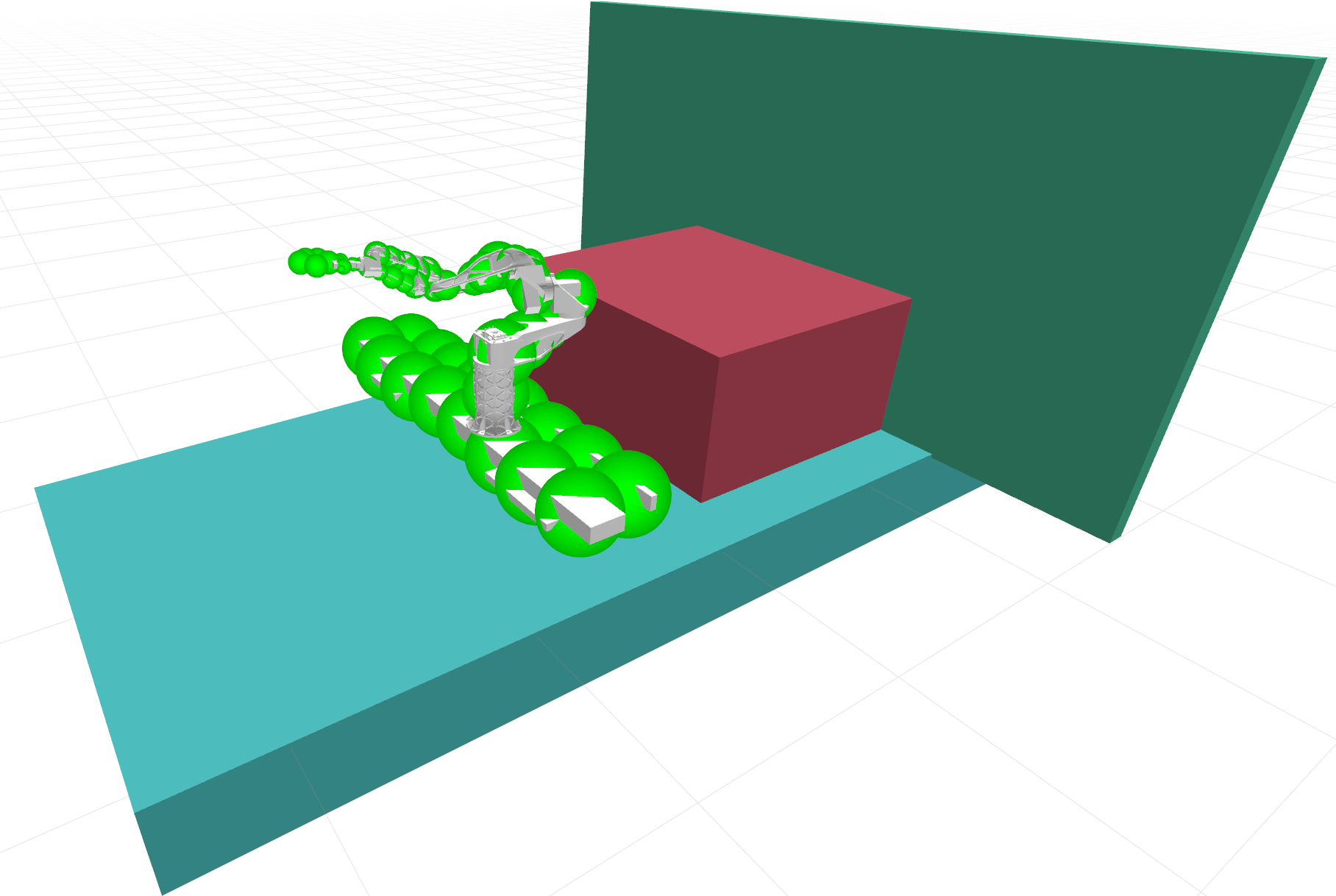}
\caption{Collision representation used for feasibility evaluation. Robot links are approximated by collision spheres and workspace obstacles by cuboids, enabling efficient collision checking along generated trajectories.}
\label{fig:robot_collision_representation}
\end{figure}

World collisions and self-collisions are evaluated explicitly along the decoded trajectory. The robot and workspace are approximated by primitive collision geometries, shown in Fig.~\ref{fig:robot_collision_representation}. During training, these collisions contribute a differentiable collision loss following \cite{sundaralingam2023curoboparallelizedcollisionfreeminimumjerk}.

\subsection{Trajectory Representation}
\label{subsec:traj_rep}

GASP predicts the free interior control points of a clamped B-spline defined over normalized time $\tau \in [0,1]$. The physical execution duration is the horizon $T$, with

\begin{equation}
\label{eq:time_scaling}
    t = T\tau.
\end{equation}

\subsubsection{B-spline trajectory parameterization}
\label{subsec:bspline_param}

We parameterize the joint position trajectory using a clamped B-spline of degree $d=7$. Let the total number of control points be $n_c$, with $\mathcal{C} = \{\mathbf{c}_0, \mathbf{c}_1, \ldots, \mathbf{c}_{n_c-1}\},\, \mathbf{c}_i \in \mathbb{R}^n$. The normalized-time trajectory is evaluated as

\begin{equation}
\label{eq:bspline}
\mathbf{q}(\tau)
=
\sum_{i=0}^{n_c-1} N_{i,d}(\tau)\mathbf{c}_i,
\qquad
\tau \in [0,1],
\end{equation}

where $N_{i,d}$ is the B-spline basis function for control point $\mathbf{c}_i$. The network predicts only the unconstrained interior control points $\mathcal{C}_{\mathrm{free}}$; using Eq.~\ref{eq:boundary_states}, the first three and last two control points are computed analytically as derived in \cite{kicki2023fastkinodynamicplanningconstraint}. The full control point sequence is

\begin{equation}
    \label{eq:learnable_points}
    \mathcal{C}
    =
    \{\mathbf{c}_0, \mathbf{c}_1, \mathbf{c}_2\}
    \cup
    \mathcal{C}_{\mathrm{free}}
    \cup
    \{\mathbf{c}_{n_c-2}, \mathbf{c}_{n_c-1}\},
\end{equation}

where $\mathcal{C}_{\mathrm{free}}$ contains the free control points predicted by the model. In the residual formulation, the model predicts offsets relative to baseline free control points:

\begin{equation}
\label{eq:residual_cp}
\mathcal{C}_{\mathrm{free}}
=
\mathcal{C}_{\mathrm{baseline}}
+
\Delta \mathcal{C}_{\theta},
\end{equation}

where $\mathcal{C}_{\mathrm{baseline}}$ is obtained by interpolation and $\Delta \mathcal{C}_{\theta}$ is the network-predicted residual. We use the analytical boundary construction of \cite{kicki2023fastkinodynamicplanningconstraint}; Appendix~\ref{app:bspline_precompute} summarizes the matrix precomputation used for fast inference.

\subsection{Model architecture}
\label{sec:nmp_arch}

The planner is implemented as a CNN over the temporal sequence of B-spline control points. It maps boundary conditions to residual corrections of the baseline trajectory in Eq.~\ref{eq:residual_cp}, improving feasibility and smoothness while leaving analytically constrained boundary points fixed. Fig.~\ref{fig:pipeline} summarizes the architecture.

\subsubsection{Input encoding and normalization}

The network input concatenates the initial and final boundary states: position, velocity, and acceleration at $t=0$, and position and velocity at $t=T$. Boundary quantities are normalized by their physical limits for numerical conditioning. Given lower and upper limits $\mathbf{\ell}$ and $\mathbf{u}$, we use
$\bar{\textbf{q}} = \textbf{q}_{\text{norm}} = 2 \frac{\textbf{q} - (\mathbf{\ell} + \mathbf{u})/2}{\mathbf{u} - \mathbf{\ell}},$
yielding values approximately in $[-1,1]$. In Fig.~\ref{fig:pipeline} the normalized inputs are denoted as $\bar{\zeta}(\cdot)$.

\subsubsection{Baseline trajectory construction}

Before neural refinement, GASP constructs a baseline trajectory in control-point space. Boundary control points are computed in closed form to enforce the prescribed derivatives, and the remaining internal control points are initialized by linear interpolation. This baseline satisfies the boundary constraints, B-spline smoothness, and consistency with the horizon $T$.

\subsubsection{Latent sampling layer}
\label{sec:latent_sampling}
To generate multiple candidates per query, GASP uses a latent variable $\mathbf{z} \in \mathbb{R}^{d_z}$. During training, an encoder maps the normalized boundary input, optionally with the baseline free control points, to a Gaussian posterior with mean $\boldsymbol{\mu}_q$ and diagonal covariance $\Sigma_q = \mathrm{diag}(\boldsymbol{\sigma}_q^2)$. Samples are drawn via $\mathbf{z} = \boldsymbol{\mu}_q + \boldsymbol{\sigma}_q \odot \boldsymbol{\epsilon}$, with $\boldsymbol{\epsilon} \sim \mathcal{N}(0,I)$, and regularized toward a standard normal prior by a KL term. The latent vector is mapped to control-point offsets before convolutional refinement. At inference, $\mathbf{z}$ is sampled from the prior, allowing multiple trajectory proposals to be decoded and validated in parallel.

\subsubsection{Convolutional refinement network}

The core planner is a one-dimensional CNN over the temporal trajectory dimension. Its input sequence is formed by sampling the current B-spline trajectory and concatenating baseline control points, positions $\mathbf{q}(t)$, velocities $\dot{\mathbf{q}}(t)$, accelerations $\ddot{\mathbf{q}}(t)$, and jerks $\dddot{\mathbf{q}}(t)$. The CNN uses 1D convolutions, nonlinearities, and skip connections to predict residual corrections only for the unconstrained midpoint region; boundary control points remain analytically fixed.
\begin{figure}[t]
    \centering
    \input{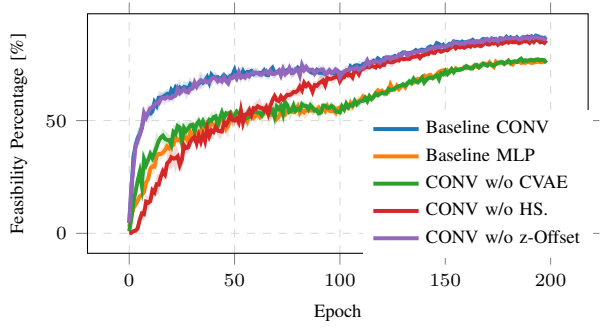}
    \caption{
    Training evolution of the feasibility percentage for the proposed and ablated architectures over five runs ($n=5$). Solid curves show the mean feasibility over epochs, while shaded regions indicate the corresponding variability bounds.
    }
    \label{fig:feasibility_training}
\end{figure}
\subsection{Model Training}
\label{subsec:model_training}

During training, decoded trajectories as defined in Eq.~\ref{eq:trajectory_map} are optimized using differentiable penalties corresponding to the planner's feasibility and regularization objectives. We write the training objective as

\begin{equation}
\label{eq:training_objective}
\begin{aligned}
\mathcal{L}_{\mathrm{train}}
&=
\lambda_{\mathrm{kin}} \mathcal{L}_{\mathrm{kin}}
+ \lambda_{\mathrm{col}} \mathcal{L}_{\mathrm{col}}
+ \lambda_{\mathrm{jerk}} \mathcal{L}_{\mathrm{jerk}} \\
&\quad
+ \lambda_{\mathrm{time}} \mathcal{L}_{\mathrm{time}}
+ \lambda_{\mathrm{KL}} \mathcal{L}_{\mathrm{KL}}.
\end{aligned}
\end{equation}

Here, $\mathcal{L}_{\mathrm{kin}}$ penalizes violations of the position, velocity, acceleration, and jerk limits in Eq.~\ref{eq:bounds}; $\mathcal{L}_{\mathrm{col}}$ penalizes world and self-collisions; $\mathcal{L}_{\mathrm{jerk}}$ regularizes smoothness; $\mathcal{L}_{\mathrm{time}}$ encourages time-efficient motion; and $\mathcal{L}_{\mathrm{KL}}$ regularizes the latent module from Sec.~\ref{sec:latent_sampling}. The loss components are detailed in Appendix~\ref{app:lossfunctions}.

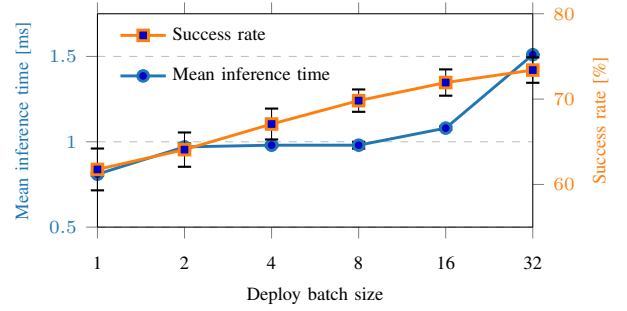
\begin{figure}[t]
    \centering
    \begin{tikzpicture}

\begin{axis}[
	width=0.85\linewidth,
	height=0.51\linewidth,
	xmode=log,
	log basis x={2},
	xmin=1,
	xmax=32,
	ymin=0.50,
	ymax=1.75,
	xlabel={Deploy batch size},
	ylabel={Mean inference time [ms]},
	xtick={1,2,4,8,16,32},
	xticklabels={{1},{2},{4},{8},{16},{32}},
	grid=major,
	major grid style={dashed,gray!50},
	tick align=outside,
	xmajorgrids=false,
	ymajorgrids=true,
	clip=false,
	axis y line*=left,
	axis x line*=bottom,
	legend style={
		draw=none,
		font=\scriptsize,
		at={(0.03,0.8)},
		anchor=north west
	},
    xlabel style={font=\scriptsize},
    ylabel style={font=\scriptsize, color={rgb,255:red,31;green,119;blue,180}},
    x tick label style={font=\scriptsize, yshift=-3pt},
    y tick label style={font=\scriptsize, text={rgb,255:red,31;green,119;blue,180}},
]
\addplot+[
	color={rgb,255:red,31;green,119;blue,180},
	mark=*,
	line width=1.10pt,
	error bars/y dir=both,
	error bars/y explicit,
	error bars/error bar style={black},
	error bars/error mark=-,
	error bars/error mark options={black, rotate=90, mark size=2.5pt, line width=0.8pt}
] coordinates {
	(1.0,0.81) +- (0,0.01)
	(2.0,0.97) +- (0,0.02)
	(4.0,0.98) +- (0,0.01)
	(8.0,0.98) +- (0,0.02)
	(16.0,1.08) +- (0,0.01)
	(32.0,1.51) +- (0,0.00)
};
\addlegendentry{Mean inference time}
\end{axis}

\begin{axis}[
	width=0.85\linewidth,
	height=0.51\linewidth,
	xmode=log,
	log basis x={2},
	xmin=1,
	xmax=32,
	ymin=55.00,
	ymax=80.00,
	xtick={1,2,4,8,16,32},
	xticklabels=\empty,
	ylabel={Success rate [\%]},
	tick align=outside,
	clip=false,
	axis y line*=right,
	axis x line*=top,
	x tick label style={draw=none},
	legend style={
		draw=none,
		font=\scriptsize,
		at={(0.03,0.99)},
		anchor=north west
	},
    ylabel style={font=\scriptsize, color={rgb,255:red,255;green,127;blue,14}},
    y tick label style={font=\scriptsize, text={rgb,255:red,255;green,127;blue,14}},
]
\addplot+[
	color={rgb,255:red,255;green,127;blue,14},
	mark=square*,
	line width=1.10pt,
	error bars/y dir=both,
	error bars/y explicit,
	error bars/error bar style={black},
	error bars/error mark=-,
	error bars/error mark options={black, rotate=90, mark size=2.5pt, line width=0.8pt}
] coordinates {
	(1.0,61.76) +- (0,2.45)
	(2.0,64.08) +- (0,2.01)
	(4.0,67.08) +- (0,1.81)
	(8.0,69.82) +- (0,1.31)
	(16.0,71.94) +- (0,1.54)
	(32.0,73.40) +- (0,1.49)
};
\addlegendentry{Success rate}
\end{axis}

\end{tikzpicture}
    \caption{Success rate of the convolutional planner as a function of deployment batch size, evaluated on a dataset with (i)~random non-stationary initial states, (ii)~random stationary final states, and (iii)~analytical-solution filtering for limit feasibility.
    }
    \label{fig:deploy_batch_size}
\end{figure}
\subsection{Data Generation}
\label{subsec:data_generation}

GASP is trained from boundary-condition pairs rather than expert trajectories. Training samples are produced online by an on-GPU sampler, while validation uses fixed offline datasets.

Each sample consists of Eq.~\ref{eq:boundary_states} and a horizon $T$, corresponding at the position level to $(\mathbf{q}_0, \mathbf{q}_T, T)$. Initial positions are sampled collision-free, and initial velocities and accelerations are sampled within limits unless forced to zero. Two on-GPU rejection filters keep samples well posed for the current horizon: a \emph{reachability-aware} filter restricts initial positions to the velocity- and acceleration-reachable region around the goal under a per-joint bang-bang model, intersected with hard joint limits; and an \emph{inevitable-violation} filter rejects samples whose short-horizon forward propagation would unavoidably violate position limits or collide within $0.1\,\mathrm{s}$. Rejected samples are resampled in place on the GPU.

Offline validation datasets use the same sampler, followed by analytic-planner filtering. Samples whose minimum feasible analytic duration exceeds $T$ are rejected as kinematically infeasible. The remaining samples are partitioned into \emph{in-collision} and \emph{collision-free} groups according to whether the analytic trajectory collides, allowing us to evaluate whether GASP recovers feasible alternatives when the analytic method fails due to collisions.

For the table-tennis planner in Sec.~\ref{sec:gasp_tabletennis}, training and validation goals are sampled from three reset groups: \emph{homefinal}, the nominal home configuration with zero terminal velocity; \emph{anyfinal}, a random collision-free joint configuration with zero terminal velocity by default; and \emph{homegridfinal}, configurations from a precomputed grid of reset poses.
\begin{figure}[t]
    \centering
\input{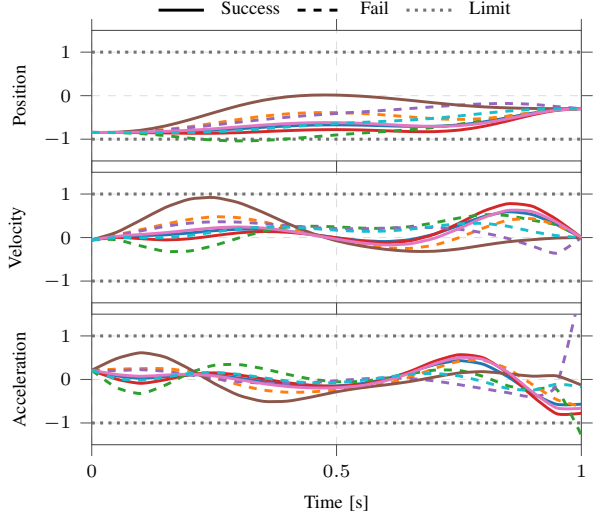}
\caption{
Normalized joint-space trajectories for representative successful and failed samples. The three stacked plots show position, velocity, and acceleration over normalized execution time. Solid and dashed lines denote successful and failed trajectories, respectively, with dotted horizontal lines marking normalized constraint limits.
}
\label{fig:trajectory_limits}
\end{figure}
\section{Experiments and Results}
We evaluate GASP along three axes: training-time feasibility, offline comparison with analytical, optimization-based, and neural baselines, and deployment as an online reset planner for reinforcement learning and robotic table tennis. Unless stated otherwise, evaluations use the default configuration in Figure~\ref{fig:pipeline}. Success denotes satisfying boundary constraints, joint position, velocity, acceleration, and jerk limits, and collision-free execution at deployment resolution. Inference time includes latent sampling, decoding, validation, and candidate selection on the GPU.

\subsection{Training}
Fig.~\ref{fig:feasibility_training} shows the evolution of the validation feasibility metric during training for the full convolutional model and three ablations that remove individual components (CVAE branch, horizon curriculum, and latent-driven midpoint offset). The full CONV model converges to the highest aggregate feasibility, and the gap to the ablated variants is consistent with the role each component plays in widening the distribution of feasible candidates per query. Fig.~\ref{fig:trajectory_limits} shows representative successful and failed trajectory candidates drawn for the same start--goal pair, illustrating both the smoothness imposed by the B-spline parameterization and the constraint margins that the planner is trying to respect.

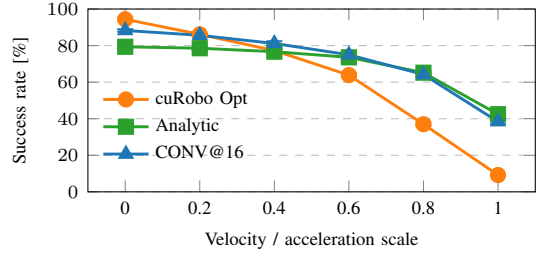
\begin{figure}[t]
\centering
\begin{tikzpicture}
\begin{axis}[
	width=7.5cm,
	height=4.cm,
	xlabel={Velocity / acceleration scale},
	ylabel={Success rate [\%]},
	ymin=0.00,
	ymax=100.00,
    legend cell align={left},
	legend style={draw=none, fill=none, at={(0.2,.1)}, anchor=south, legend columns=1,font=\scriptsize},
	major grid style={dashed,gray!50},
	ymajorgrids=true,
	xmajorgrids=false,
	xmode=normal,
	xtick={1.00,0.80,0.60,0.40,0.20,0.00},
	xticklabels={{1},{0.8},{0.6},{0.4},{0.2},{0}},
    xlabel style={font=\scriptsize},
    ylabel style={font=\scriptsize},
    x tick label style={
        font=\scriptsize,
    },
    y tick label style={font=\scriptsize},
    xmajorgrids=false,
	ymajorgrids=true,
	clip=false,
]
\addplot+[color={rgb,255:red,255;green,127;blue,14}, line width=1.10pt, mark=*, mark options={draw={rgb,255:red,255;green,127;blue,14}, fill={rgb,255:red,255;green,127;blue,14}}, mark size=2.50pt, error bars/y dir=both, error bars/y explicit, error bars/error mark=-, error bars/error mark options={rotate=90, mark size=3pt, line width=0.8pt, draw={rgb,255:red,255;green,127;blue,14}}] coordinates {(1.00,9.18) +- (0,0.46) (0.80,37.00) +- (0,1.01) (0.60,63.84) +- (0,1.03) (0.40,77.28) +- (0,1.13) (0.20,86.16) +- (0,0.46) (0.00,94.40) +- (0,1.13)};
\addlegendentry{cuRobo Opt}
\addplot+[color={rgb,255:red,44;green,160;blue,44}, line width=1.10pt, mark=square*, mark options={draw={rgb,255:red,44;green,160;blue,44}, fill={rgb,255:red,44;green,160;blue,44}}, mark size=2.50pt, error bars/y dir=both, error bars/y explicit, error bars/error mark=-, error bars/error mark options={rotate=90, mark size=3pt, line width=0.8pt, draw={rgb,255:red,44;green,160;blue,44}}] coordinates {(1.00,42.46) +- (0,1.32) (0.80,65.16) +- (0,0.69) (0.60,73.58) +- (0,0.57) (0.40,76.66) +- (0,0.89) (0.20,78.54) +- (0,1.15) (0.00,79.34) +- (0,0.98)};
\addlegendentry{Analytic}
\addplot+[color={rgb,255:red,31;green,119;blue,180}, line width=1.10pt, mark=triangle*, mark options={draw={rgb,255:red,31;green,119;blue,180}, fill={rgb,255:red,31;green,119;blue,180}}, mark size=2.50pt, error bars/y dir=both, error bars/y explicit, error bars/error mark=-, error bars/error mark options={rotate=90, mark size=3pt, line width=0.8pt, draw={rgb,255:red,31;green,119;blue,180}}] coordinates {(1.00,38.40) +- (0,1.10) (0.80,64.16) +- (0,1.13) (0.60,74.98) +- (0,1.13) (0.40,81.16) +- (0,1.19) (0.20,85.62) +- (0,0.76) (0.00,88.28) +- (0,0.66)};
\addlegendentry{CONV@16}
\end{axis}
\end{tikzpicture}
\caption{
Success rate under nonstationary initial-state conditions induced by velocity and acceleration scaling. The horizontal axis is the scaling factor, with $0$ denoting a stationary initial state and $1$ the fully nonstationary state within limits.
}
\label{fig:nonstationary_initial_state_scaling}
\end{figure}

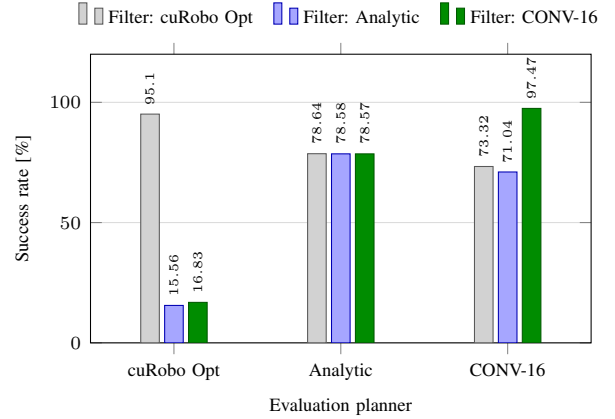
\begin{figure}[t]
\centering

\pgfplotsset{
  colormap={BuGnMedium}{
    rgb255=(247,252,253);
    rgb255=(229,245,249);
    rgb255=(204,236,230);
    rgb255=(153,216,201);
    rgb255=(102,194,164);
    rgb255=(65,174,118);
    rgb255=(40,150,85);
    rgb255=(20,125,65);
  }
}

\begin{tikzpicture}
\begin{axis}[
    ybar,
    width=0.95\columnwidth,
    height=5.4cm,
    bar width=7pt,
    ymin=0,
    ymax=120,
    ylabel={Success rate [\%]},
    xlabel={Evaluation planner},
    symbolic x coords={cuRobo Opt,Analytic,CONV-16},
    xtick=data,
    x tick label style={
        font=\scriptsize,
        rotate=0,
        align=right
    },
    y tick label style={font=\scriptsize},
    label style={font=\scriptsize},
    enlarge x limits=0.25,
    ymajorgrids=true,
    grid style={line width=0.2pt, draw=gray!30},
    legend style={
        font=\scriptsize,
        at={(0.5,+1.20)},
        anchor=north,
        legend columns=3,
        draw=none,
        /tikz/every even column/.append style={column sep=0.2cm}
    },
    nodes near coords,
    every node near coord/.append style={
        font=\tiny,
        rotate=90,
        anchor=west
    },
]
\legend{
    Filter: cuRobo Opt,
    Filter: Analytic,
    Filter: CONV-16
}
\addplot[
    fill=gray!35,
    draw=gray!70!black
] coordinates {
    (cuRobo Opt,95.10)
    (Analytic,78.64)
    (CONV-16,73.32)
};

\addplot[
    fill=blue!35,
    draw=blue!70!black
] coordinates {
    (cuRobo Opt,15.56)
    (Analytic,78.58)
    (CONV-16,71.04)
};

\addplot[
    fill=green!55!black,
    draw=green!35!black
] coordinates {
    (cuRobo Opt,16.83)
    (Analytic,78.57)
    (CONV-16,97.47)
};

\end{axis}
\end{tikzpicture}
\caption{Success rate grouped by evaluation planner. Each group corresponds to the evaluation planner, and the bars within each group correspond to the planner used to filter the dataset.
}
\label{fig:compare_filtering_success_heatmap}
\end{figure}

\begin{table*}[t]
\centering
\caption{Detailed failure breakdown across planner types on the shared evaluation set.}
\label{tab:compare-planner-types-failures-ruckigfiltered}
\begin{tabular}{lccccc}
\hline
Failure mode & CONV@16 & CONV@1 & MLP & cuRobo Opt & Analytic \\
\hline
Collision rate [\%] & 4.28 $\pm$ 0.83 & 7.72 $\pm$ 2.00 & \textbf{3.84 $\pm$ 0.54} & 83.94 $\pm$ 1.29 & 20.88 $\pm$ 0.91 \\
Position-limit violation [\%] & \textbf{3.82 $\pm$ 0.38} & 4.50 $\pm$ 0.43 & 5.20 $\pm$ 0.44 & 83.12 $\pm$ 1.26 & {0.00 $\pm$ 0.00} \\
Velocity-limit violation [\%] & 1\textbf{8.18 $\pm$ 1.56} & 19.88 $\pm$ 1.41 & 19.74 $\pm$ 1.54 & 83.12 $\pm$ 1.26 & {0.00 $\pm$ 0.00} \\
Acceleration-limit violation [\%] & \textbf{9.20 $\pm$ 0.92} & 11.76 $\pm$ 1.31 & 9.76 $\pm$ 0.66 & 83.12 $\pm$ 1.26 & {0.00 $\pm$ 0.00} \\
Jerk-limit violation [\%] & \textbf{18.88 $\pm$ 1.29} & 23.64 $\pm$ 1.33 & 19.94 $\pm$ 1.55 & 83.12 $\pm$ 1.26 & {0.00 $\pm$ 0.00} \\
\hline
Success rate [\%] & \textbf{71.94 $\pm$ 1.54} & 61.76 $\pm$ 2.45 & 67.00 $\pm$ 1.49 & 16.06 $\pm$ 1.29 & {79.12 $\pm$ 0.91} \\
Inference Time [ms] & 1.08 $\pm$ 0.01 & 0.82 $\pm$ 0.02 & \textbf{0.77 $\pm$ 0.03} & 56.78 $\pm$  0.73 & {4.24 $\pm$ 0.03} \\
\hline
\end{tabular}

\end{table*}

\subsection{Experiments}
We evaluate GASP along three axes that match the contributions claimed in the introduction. First, we compare against state-of-the-art planners (Sec.~\ref{sec:exp_sota}). Second, we ablate the main components of the architecture and the training schedule using the tables in this section. Third, we report the use of GASP as an online reset planner inside an RL pipeline and as a real-time motion-generation component for a dynamic robot task (Sec.~\ref{sec:gasp_tabletennis}).

\subsubsection{Comparison with State of the Art}
\label{sec:exp_sota}
We compare GASP against (i)~Ruckig~\cite{berscheid2021jerk}, an analytic real-time jerk-limited online trajectory generator that does not perform collision checking, (ii)~the cuRobo optimization-based motion planner, and (iii)~a neural planner using an MLP architecture, following the method proposed in \cite{kicki2022speeding}. The CONV variant of GASP with deployment batch size~$16$ (CONV-16) is evaluated against both baselines on the shared evaluation set used in Table~\ref{tab:compare_model_ablations_ruckigfiltered_compact}. Detailed per-failure-mode breakdowns are reported in Table~\ref{tab:compare-planner-types-failures-ruckigfiltered}, and Fig.~\ref{fig:nonstationary_initial_state_scaling} shows the impact of nonzero initial velocity and acceleration on the success rate where the proposed CONV@16 planner maintains performance comparable to the analytic baseline over moderate nonstationarity and substantially outperforms cuRobo optimization at higher scaling factors, indicating improved robustness to nonzero initial velocity and acceleration. Fig.~\ref{fig:compare_filtering_success_heatmap} reports a planner-compatibility view in which evaluation and filtering planners are varied independently where CONV@16 remains strong across all filtered datasets, while cuRobo Opt performs well only on data filtered by itself.

\begin{table}[t]
\centering
\caption{
Compact ablation study of the convolutional motion planner. All entries use $5$ trials. 
Parameters are reported in millions. The dataset is filtered for limit violations using the analytical solution. All experiments shown here use 150 training epochs.
}
\label{tab:compare_model_ablations_ruckigfiltered_compact}
\scriptsize
\setlength{\tabcolsep}{3pt}
\resizebox{\columnwidth}{!}{
\begin{tabular}{llccc}
\toprule
Study & Variant & P [M] & Success [\%] & Time [ms] \\
\midrule
\multirow{2}{*}{Base}
& CONV @16 & 4.26 & \textbf{67.70 $\pm$ 0.83 }& 1.08 $\pm$ 0.00 \\
& CONV @1  & 4.26 & 58.20 $\pm$ 0.88 &\textbf{0.80 $\pm$ 0.00} \\

\midrule
\multirow{6}{*}{CVAE}
& w/o CVAE @16       & 4.09 & 59.88 $\pm$ 1.11 & 1.04 $\pm$ 0.00 \\
& w/o CVAE @1        & 4.09 & 57.30 $\pm$ 1.01 & \textbf{0.79 $\pm$ 0.01} \\
& w/o $z$ Concat @16 & 4.26 & 63.58 $\pm$ 1.44 & 1.07 $\pm$ 0.01 \\
& w/o $z$ Concat @1  & 4.26 & 51.90 $\pm$ 1.89 & 0.81 $\pm$ 0.01 \\
& w/o $z$ Offset @16 & 4.26 & 67.24 $\pm$ 1.71 & 1.05 $\pm$ 0.00 \\
& w/o $z$ Offset @1  & 4.26 & 59.94 $\pm$ 2.03 & 0.82 $\pm$ 0.01 \\

\midrule
Horizon
& w/o sched. & 4.26 & 65.38 $\pm$ 0.94 & 1.08 $\pm$ 0.01 \\

\midrule
\multirow{3}{*}{Groups}
& $g=2$ & 2.30 & 63.88 $\pm$ 1.87 & 1.10 $\pm$ 0.01 \\
& $g=4$ & 1.31 & 59.80 $\pm$ 1.75 & 0.97 $\pm$ 0.00 \\
& $g=8$ & 0.82 & 52.72 $\pm$ 2.13 & 0.96 $\pm$ 0.00 \\

\midrule
\multirow{2}{*}{Kernel}
& $k=3$ & 2.64 & 40.22 $\pm$ 20.19 & 1.02 $\pm$ 0.02 \\
& $k=7$ & 5.89 & 67.12 $\pm$ 1.33 & 1.11 $\pm$ 0.00 \\
\bottomrule
\end{tabular}
}
\end{table}

\subsection{Ablation Studies}
We ablate the main components of the convolutional planner along three axes: (i)~the conditional latent module, (ii)~the horizon-curriculum schedule, and (iii)~the convolutional capacity (group count and kernel size). 

Table~\ref{tab:compare_model_ablations_ruckigfiltered_compact} summarizes the ablation studies of the convolutional motion planner. The \textit{Baseline CONV} model denotes the full model with the CVAE module. We compare it against a deterministic variant \textit{w/o CVAE}, which disables the CVAE module entirely; a variant \textit{w/o $z$ Concat}, which keeps the CVAE but does not concatenate the latent code $z$ to the convolutional inputs; and a variant \textit{w/o $z$ Offset}, which removes the latent-derived offset added to the initial linearly interpolated free control points. The table also reports the effect of removing the default $1.5\,\mathrm{s}\!\to\!1.0\,\mathrm{s}$ horizon-curriculum schedule and training directly with a fixed $1.0\,\mathrm{s}$ horizon. Finally, it sweeps the number of convolutional groups in the hidden convolutional layers, from the default $g=1$ without grouping to $g\in\{2,4,8\}$, and compares different convolutional kernel sizes while keeping all other hyperparameters fixed. For all entries, the evaluation dataset is filtered for limit violations using the analytical solution, so that samples that are inevitable to violate limits are excluded.

Table~\ref{tab:compare_planner_type_ablations_ruckigfiltered_compact} compares the three deployment back-ends. The table reports the success rate, number of trainable parameters, and mean inference time for each implementation. All entries are evaluated over $5$ trials and use the same evaluation set, filtered for limit feasibility through the analytical reference solution so that samples that are kinematically impossible to satisfy are excluded.

\begin{table}[t]
\centering
\caption{
Planner implementation comparison. All entries use $5$ trials. Parameters are reported in millions. 
The dataset is filtered for limit violations using the analytical solution.
}
\label{tab:compare_planner_type_ablations_ruckigfiltered_compact}
\scriptsize
\setlength{\tabcolsep}{3pt}
\resizebox{\columnwidth}{!}{
\begin{tabular}{llccc}
\toprule
Study & Variant & P [M] & Success [\%] & Time [ms] \\
\midrule
\multirow{3}{*}{CONV @16}
& \texttt{default}       & 4.26  & \textbf{71.94 $\pm$ 1.54} & 1.05 $\pm$ 0.00 \\
& \texttt{CPU-ONNX}      & 4.26  & 57.26 $\pm$ 2.08 & 3.94 $\pm$ 0.09 \\
& \texttt{w/o coll. check} & 4.26  & 67.26 $\pm$ 2.18 & 0.75 $\pm$ 0.00 \\

\midrule
\multirow{3}{*}{CONV @1}
& \texttt{default}       & 4.26  & 61.76 $\pm$ 2.45 & 0.80 $\pm$ 0.01 \\
& \texttt{CPU-ONNX}      & 4.26  & 61.76 $\pm$ 2.45 & 2.75 $\pm$ 0.05 \\
& \texttt{w/o coll. check} & 4.26  & 61.76 $\pm$ 2.45 & 0.62 $\pm$ 0.02 \\

\midrule
\multirow{3}{*}{MLP}
& \texttt{default}       & 17.00 & 67.00 $\pm$ 1.49 & 0.74 $\pm$ 0.00 \\
& \texttt{CPU-ONNX}      & 17.00 & 67.00 $\pm$ 1.49 & 1.85 $\pm$ 0.03 \\
& \texttt{w/o coll. check} & 17.00 & 67.00 $\pm$ 1.49 & \textbf{0.60 $\pm$ 0.01} \\
\bottomrule
\end{tabular}
}
\end{table}

\subsection{GASP in Real-World Table Tennis}
\label{sec:gasp_tabletennis}
\begin{figure}[t]
    \centering
    \includegraphics[width=0.8\linewidth]{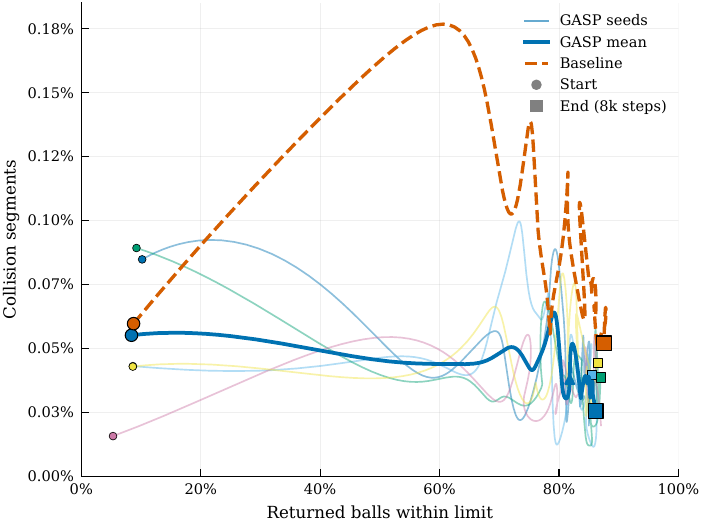}
    \caption{
    Collision reduction during training. Curves show return rate within the time limit versus the fraction of colliding trajectory segments over 8,000 training steps. The lower-right region is preferred, indicating high return performance and low collision frequency. Thin curves denote GASP seeds ($n{=}5$), the solid curve the GASP mean, and the dashed curve the Baseline; circles and squares mark the start and end of training.
}
    \label{fig:policy_collision_return}
\end{figure}

We evaluate real-world deployment on the table-tennis system of~\cite{Durr2026Ace}, using the same robotic platform and focusing on collision-aware reset motion generation. GASP serves as the reset planner for a multi-step reinforcement-learning agent: at each episode start, the agent issues a start--goal query from the current robot state to the desired reset configuration. The query is replicated across a deployment batch of $16$ by default, and trajectory candidates are decoded and validated in parallel on the GPU. The selector returns the first feasible candidate under the configured rule, using the speed heuristic by default. To avoid inference latency spikes, the GPU-resident model is kept warm with high-frequency dummy planning calls before each episode. Fig.~\ref{fig:policy_collision_return} shows that GASP improves collision avoidance without degrading task performance: it reaches a return rate comparable to the Baseline ($0.858$ vs.\ $0.873$), while reducing colliding trajectory segments by about $50\%$ during training.

We further deploy a policy trained with GASP as the reset planner in a real-time robotic table-tennis system against an elite-level player. GASP runs at the robot control frequency on the same GPU as the perception and policy stack, using the deployment-resolution decoder with $\Delta t = 0.001\mathrm{s}$. For each new tactical target, the planner generates a coupled joint-space reset trajectory that satisfies per-joint kinematic and dynamic limits and remains collision-free with respect to the table-tennis environment. Videos of the real-system deployment are provided in the supplementary materials.

\section{Discussion}
GASP is built around three design choices introduced in Sec.~\ref{sec:method}: a boundary-conditioned clamped B-spline trajectory representation, a lightweight convolutional residual network that predicts only the free interior control points, and a sampleable latent variable that allows multiple candidate trajectories to be generated for the same start--goal query. The experiments show that GASP’s representation, architecture, and sampling procedure are mutually reinforcing. GASP satisfies the prescribed start and goal derivatives by construction and concentrates learning on the interior motion, where most feasibility failures occur.
The ablation results further show that the full combination of convolutional feature extraction, latent sampling, and curriculum-based training provides the highest feasibility, supporting GASP’s formulation as a batched, sampleable planner rather than a deterministic trajectory regressor.

Across the shared evaluation sets, GASP remains competitive with the analytical baseline while adding explicit collision checking, and it is more robust than the GPU optimization baseline when the initial state contains nonzero velocity and acceleration. The failure breakdowns further show that residual errors are concentrated in the interior trajectory region rather than at the endpoints, consistent with enforcing boundary conditions analytically.

The sampleable convolutional model benefits from batched candidate generation. Because latent sampling, decoding, limit checking, and collision checking are all GPU-parallelized, larger deployment batches improve success rate with approximately constant per-batch latency, as shown in Fig.~\ref{fig:deploy_batch_size}. Compared with the analytical solution, GASP adds trajectory-level collision avoidance and coupled joint-space planning while maintaining a comparable inference-time envelope on GPU. Compared with the cuRobo optimization baseline, GASP has substantially lower inference-time variance and is more robust to nonzero initial velocity and acceleration (Fig.~\ref{fig:nonstationary_initial_state_scaling}), at the cost of robot-specific training.

GASP remains constrained by training that is specific to each robot, tool, and kinematic-limit setting, by collision representations based on static primitive shapes, and by manually tuned fixed loss weights. These limitations suggest future work on scale-aware adaptation, dynamic collision modeling, and learned objective balancing.
\section{Conclusion}
We presented GASP, a GPU-accelerated, sampleable, and collision-aware joint-space motion planner in real-time. The method combines a residual clamped B-spline parameterization with a convolutional residual neural predictor and a conditional latent variable, allowing multiple trajectory candidates to be generated, decoded, and validated in parallel on the GPU for the same start--goal query. The training objective is feasibility-driven: it combines differentiable joint-limit penalties on position, velocity, acceleration, and jerk with world- and self-collision penalties and a smoothness/time-optimality regularizer. Empirically, the resulting planner is competitive with established analytical and optimization-based baselines on the evaluated robot, while exposing a sampleable interface that can be naturally embedded in higher-level learning and control loops. Future work will focus on multi-robot generalization, richer obstacle representations, and learned objective-weight scheduling.

\bibliographystyle{IEEEtran}
\bibliography{references}

\appendix

\subsection{Methods}
\label{app:methods}

\subsubsection{B-Spline Precomputation}
\label{app:bspline_precompute}

For fast dense decoding, GASP precomputes and caches the clamped B-spline basis matrices for position, velocity, acceleration, and jerk, corresponding to $N$, $\partial N/\partial \tau$, $\partial^2 N/\partial \tau^2$, and $\partial^3 N/\partial \tau^3$. Trajectories are then decoded by batched matrix multiplication. Derivatives are converted from normalized to physical time using Eq.~\ref{eq:time_scaling}: $\dot{\mathbf q}(t)=T^{-1}\partial \mathbf q/\partial \tau$, $\ddot{\mathbf q}(t)=T^{-2}\partial^2 \mathbf q/\partial \tau^2$, and $\dddot{\mathbf q}(t)=T^{-3}\partial^3 \mathbf q/\partial \tau^3$.

\subsubsection{Model Training}
\label{app:nmp_training}

Training uses PyTorch Lightning with AdamW, learning rate $5\times10^{-4}$, weight decay $10^{-4}$, $\beta=(0.9,0.98)$, AMSGrad, gradient clipping at norm~$1$, and cosine annealing over $600$ epochs. The default minibatch size is $64$; because samples are generated online, each epoch is defined as $256$ optimizer steps, with validation every epoch on four batches of size $256$. A horizon curriculum anneals $T$ from $1.5\,\mathrm{s}$ to $1.0\,\mathrm{s}$ over the first $300$ epochs, and the online sampler uses the same active horizon. Model selection uses the aggregate validation feasible percentage computed from the hard joint-limit and collision predicates.

\subsubsection{Inference and Candidate Selection}
\label{app:nmp_inference}

At inference, each start--goal query is replicated across a deployment batch of size $16$ by default. Latent samples are decoded, checked against joint limits and collisions in parallel on the GPU, and passed to a selector that returns a feasible candidate using the speed heuristic by default. The first candidate uses $z=\mathbf{0}$ for deterministic behavior.

\subsubsection{Loss Functions}
\label{app:lossfunctions}

We denote the decoded discrete trajectory by $\zeta = \{\mathbf{q}(t_k), \dot{\mathbf{q}}(t_k), \ddot{\mathbf{q}}(t_k), \dddot{\mathbf{q}}(t_k)\}_{k=1}^{N}$. Several losses are evaluated only on the learnable interior part of the spline, namely between $\mathbf{c}_2$ and $\mathbf{c}_{n_c-2}$ as defined in \ref{eq:learnable_points}. We denote the corresponding sample indices by $\mathcal{K}_{in}=\{k:t_k\in[t_{c_2},t_{n_{c-2}}]\}$.

Additionally, let $\tilde{.}$ denote the trajectory normalized by each joint's limits, such that all joint states are in $[-1,1]$. The losses below correspond to the terms in Eq.~\ref{eq:training_objective}.
The active stable configuration uses fixed weights for feasibility and smoothness losses, while the KL term is weighted separately by the annealing schedule in Section~\ref{sec:cvae_sched}. Specifically, $\lambda_{\mathrm{jerk}}=1000$ for jerk regularization, $\lambda_{\mathrm{col}}=100$ for world and self-collision, $\lambda_{\mathrm{pos}}=\lambda_{\mathrm{vel}}=30$ and $\lambda_{\mathrm{acc}}=60$ for kinematic bounds, $\lambda_{\mathrm{jerk\text{-}bound}}=30$, and $\lambda_{\mathrm{time}}=1$. Bound losses use cuRobo bound-cost operators with per-joint normalization to $[-1,1]$; position-bound and collision losses are masked to the learnable interior spline region.

For the kinematic limits, we use one penalty per derivative order $c\in\{p,v,a,j\}$, corresponding to position, velocity, acceleration, and jerk. 
Let $\tilde{\mathbf q}^{(c)}(t_k)\in\mathbb{R}^{n}$ collect the normalized joint values for channel $c$, 
and let $[x]_+=\max(0,x)$. Then
\begin{equation}
    \label{eq:kin_loss}
    \mathcal{L}_{\mathrm{kin}}
    =
    \sum_{c\in\{p,v,a,j\}}
    \lambda_{\mathrm{kin}}^{(c)}
    \max_{k\in\mathcal{K}_{in}}
    \bigl\|
    [|\tilde{\mathbf q}^{(c)}(t_k)|-\bar\delta^{(c)}]_+
    \bigr\|_2^2 .
\end{equation}

This applies a worst-case interior violation penalty for each derivative order. For collision avoidance, we use

\begin{equation}
    \label{eq:col_loss}
    \begin{aligned}
    \mathcal{L}_{\mathrm{col}}
    &= \max_{k\in\mathcal{K}_{in}}
    \Bigl(
        d_{\mathrm{w}}(\mathbf{q}(t_k)) + d_{\mathrm{s}}(\mathbf{q}(t_k))
    \Bigr).
    \end{aligned}
\end{equation}
Here, $d_{\mathrm{w}}$ and $d_{\mathrm{s}}$ are the differentiable world- and self-collision violation signals used by cuRobo \cite{sundaralingam2023curoboparallelizedcollisionfreeminimumjerk}. Both terms are nonnegative, vanish when the activation margin is respected, and increase with penetration depth. Taking the maximum over $\mathcal{K}_{in}$ makes a single colliding interior sample dominate the per-trajectory penalty.

The smoothness regularizer penalizes normalized jerk over the interior trajectory:
\begin{equation}
    \label{eq:jerk_loss}
    \begin{aligned}
    \mathcal{L}_{\mathrm{jerk}}
    &= \frac{1}{\lvert \mathcal{K}_{in} \rvert}\sum_{k\in\mathcal{K}_{in}}
        \Bigl\lVert
            \tilde{\dddot{\mathbf{q}}}(t_k)
        \Bigr\rVert_{2}.
    \end{aligned}
\end{equation}
Here $\tilde{\dddot{\mathbf{q}}}(t_k)$ denotes the componentwise jerk normalized by the corresponding joint jerk limits before aggregation.

The settling-time objective distinguishes between feasible and infeasible decoded trajectories:
\begin{equation}
    \label{eq:time_loss}
    \begin{aligned}
    \mathcal{L}_{\mathrm{time}}
    &= \phi\,\widetilde{\mathcal{L}}_{\mathrm{time}} + (1-\phi)\,(1 + d_{\mathrm{viol}}).
    \end{aligned}
\end{equation}
with $\widetilde{\mathcal{L}}_{\mathrm{time}}
    = \frac{1}{N_{\mathrm{l}}}
      \sum_{k=N-N_{\mathrm{l}}+1}^{N}
      \bigl\lVert \mathbf{q}(t_k) - \mathbf{q}(t_N) \bigr\rVert_{2},$ where $\phi\in\{0,1\}$ is the hard feasibility flag, equal to $1$ if and only if all boundary, joint-limit, and collision constraints are satisfied along $\zeta$, and $N_{\mathrm{l}}=\max(1,\lfloor\rho(N-1)\rfloor)$ with $\rho=0.66$. The infeasibility score is $d_{\mathrm{viol}}
    = \max\bigl\{
        \mathcal{L}_{\mathrm{col}},
        \mathcal{L}_{\mathrm{kin}}^{(p)},
        \mathcal{L}_{\mathrm{kin}}^{(v)},
        \mathcal{L}_{\mathrm{kin}}^{(a)},
        \mathcal{L}_{\mathrm{kin}}^{(j)}
    \bigr\},$ so infeasible candidates incur a fixed offset and are then ranked by their largest collision or kinematic violation, while feasible candidates are compared through their average terminal settling behavior.

Finally, the latent regularizer follows the CVAE formulation introduced in Sec.~\ref{sec:latent_sampling}. If $q(z\mid x)$ denotes the approximate posterior conditioned on the network input $x$, we write
\begin{equation}
    \label{eq:kl_loss}
    \mathcal{L}_{\mathrm{KL}}
    = D_{\mathrm{KL}}\!\bigl(q(z \mid x)\,\Vert\,\mathcal{N}(0,I)\bigr).
\end{equation}

\subsubsection{CVAE training objective}
\label{sec:cvae_sched}

The KL weight is annealed linearly from $10^{-4}$ to $10^{-2}$ over the first $50$ epochs. With the default setting of four latent samples per training example, the code evaluates multiple stochastic forward passes and aggregates them with a softmin-style weighting controlled by a temperature parameter. This encourages the model to allocate probability mass to feasible and low-cost trajectory modes rather than to a single deterministic solution.

\end{document}